# SwinMamba: A hybrid local–global mamba framework for enhancing semantic segmentation of remotely sensed images

Qinfeng Zhu[a,b], Han Li[c], Liang He[d], and Lei Fan[a,e,1]

[a] Department of Civil Engineering, Xi'an Jiaotong-Liverpool University, Suzhou, 215123, China

[b] Department of Computer Science, University of Liverpool, Liverpool, L69 3BX, UK

[c] College of Computing and Data Science, Nanyang Technological University, 639798, Singarpore

[d] SJTUSME-COSMOPlat Joint Research Center for New Generation Industrial Intelligent Technology, Shanghai, 200240, China

[e] Design School Intelligent Built Environment Research Centre, Xi'an Jiaotong-Liverpool University, Suzhou, 215123, China

**Abstract**: Semantic segmentation of remote sensing imagery is a fundamental task in computer vision, supporting a wide range of applications such as land use classification, urban planning, and environmental monitoring. However, this task is often challenged by the high spatial resolution, complex scene structures, and diverse object scales present in remote sensing data. To address these challenges, various deep learning architectures have been proposed, including convolutional neural networks, Vision Transformers, and the recently introduced Vision Mamba. Vision Mamba features a global receptive field and low computational complexity, demonstrating both efficiency and effectiveness in image segmentation. However, its reliance on global scanning tends to overlook critical local features, such as textures and edges, which are essential for achieving accurate segmentation in remote sensing contexts. To tackle this limitation, we propose SwinMamba, a novel framework inspired by the Swin Transformer. SwinMamba integrates localized

[1] Corresponding author.

Email addresses: Lei.Fan@xjtlu.edu.cn (L. Fan)

Mamba-style scanning within shifted windows with a global receptive field, to enhance the model's perception of both local and global features. Specifically, the first two stages of SwinMamba perform local scanning to capture fine-grained details, while its subsequent two stages leverage global scanning to fuse broader contextual information. In our model, the use of overlapping shifted windows enhances inter-region information exchange, facilitating more robust feature integration across the entire image. Extensive experiments on the LoveDA and ISPRS Potsdam datasets demonstrate that SwinMamba outperforms state-of-the-art methods, underscoring its effectiveness and potential as a superior solution for semantic segmentation of remotely sensed imagery.

## 1 Introduction

Semantic segmentation of remote sensing images is a fundamental task in computer vision, with the objective of assigning each pixel in an image to predefined specific categories, such as buildings, roads, vegetation, water bodies, and bare land [1]. This pixel-level classification is of paramount importance for diverse applications, such as agricultural monitoring, urban planning, and infrastructure development [2]. Unlike semantic segmentation of natural images, remote sensing images present significant challenges stemming from their unique data characteristics [3]. These images typically exhibit high spatial resolutions [4], object scale variations, spectral inconsistencies [5], and complex spatial dependencies. Moreover, phenomena such as category overlap and blurred boundaries further challenge accurate segmentation. The aforementioned complexities far surpass the comparatively uniform scales and simpler contextual characteristics found in natural images, thereby necessitating more advanced computational strategies. Consequently, deep learning has emerged as the predominant approach to tackle this task, leveraging its capability to extract hierarchical features and efficiently process large-scale data [6].

Convolutional Neural Network (CNN)-based methods have established themselves as the mainstream approach in remote sensing semantic segmentation tasks [7], owing to their proficiency in capturing local features through convolutional operations. These methods excel at modeling the complex textures and patterns inherent in remote sensing data. However, CNNs are constrained by their limited receptive fields, which hinder their ability to capture long-range dependencies, a notable limitation when interpreting contextual information in large-scale scenes [8]. To overcome this shortfall, researchers have introduced the Transformer architecture into the visual domain [9]. Originally developed for natural language processing (NLP), the Transformer is celebrated for the superiority of its self-attention mechanism in handling contextual relationships [10]. When adapted to vision tasks, Vision Transformers (ViT) have demonstrated exceptional performance in semantic segmentation [11]. Nevertheless, their computational complexity scales quadratically with sequence length, presenting significant challenges for processing high-resolution remote sensing images. To address this bottleneck, the recently proposed Mamba model has emerged as a promising solution [12]. Also originating from the NLP domain, Mamba is designed to mitigate the quadratic complexity of Transformers while preserving a global attention-like capability. Studies have shown that Mamba exhibits remarkable adaptability in visual tasks [13], achieving outstanding results across diverse scenarios [1, 14]. In the realm of semantic segmentation of remote sensing images, numerous investigations [13, 14] have applied Vision Mamba, confirming its efficiency and effectiveness for this task.

Despite the significant progress achieved by Mamba-based semantic segmentation methods, their pervasive reliance on global scanning strategies reveals a fundamental limitation in semantic segmentation of remote sensing images where accurate delineation of spatially proximate details, such as edges and textures, are demanded. Existing Mamba-based approaches primarily focus on optimizing global scanning patterns [15], such as Z-shaped [16], S-shaped [17], diagonal [18], spiral [19], tree-based [20], or random sequences [21], to transform 2D feature maps into 1D sequences [13], often integrating dual-encoder architectures to enhance task-specific adaptability [1]. While these efforts have enhanced the model's capacity to learn diverse image patterns, these global scanning paradigms consistently treat the image as a unified global sequence [15], thereby overlooking critical local features for perception. This oversight underscores an urgent need for innovative frameworks capable of smartly integrating local and global contextual awareness, thereby paving the way for more accurate and robust pixel-level classification.

To this end, we propose SwinMamba, a novel framework enabling the fusion of local and global contextual information, making it particularly well-suited for semantic segmentation of remote sensing images. Inspired by the localized self-attention mechanisms pioneered by Swin Transformer [22], we adapt Vision Mamba's four-directional scanning paradigm to operate within local regions. This adaptation achieves a seamless synergy between the model's global modeling capabilities and local feature extraction. By incorporating an overlapping shifted window mechanism, SwinMamba facilitates more robust information exchange between local regions, mitigating the isolation issue inherent in purely local processing. To further optimize the balance between fine-grained local detail and global perception, we employ a local window-based scanning strategy in the first two stages to prioritize the capture of local details, while transitioning to global scanning in the latter two stages to acquire global contextual information. Extensive evaluations on the LoveDA and ISPRS Potsdam datasets validate the effectiveness of SwinMamba, achieving mean Intersection over Union (mIoU) improvements of 1.06% and 0.33%, respectively, over state-of-the-art Vision Mamba methods.

The main contributions of this paper are summarized as follows:

1. We introduce the SwinMamba framework featuring a novel local scanning strategy, marking the first attempt to adapt Mamba's scanning mechanism for localized regions.

2. By integrating the local scanning strategy with a shifted window mechanism, we enable effective information exchange between adjacent local regions, thereby enhancing the model's ability to integrate features across these regions.

3. We propose an integrated scanning approach that employs local scanning in initial stages and global scanning in later stages, significantly improving the fusion of local and global features.

The rest of this article is structured as follows. Section 2 reviews related works on local-global methods for segmentation and Mamba-based segmentation networks. Section 3 presents the proposed SwinMamba architecture, including its overall framework, Local SwinMamba Block, and Global SwinMamba Block. Section 4 details the experimental setup, datasets, pretraining, results, and ablation studies. Section 5 discusses the findings and outlines future work. Finally, Section 6 concludes the paper.

## 2 Related work

### 2.1 Local–global methods for segmentation

Semantic segmentation is a fundamental task in computer vision, where neural network architectures such as CNNs and ViTs each exhibit distinct strengths and weaknesses. CNNs excel in local region processing, effectively capturing fine-grained details through convolutional operations, whereas ViTs demonstrate superior proficiency in holistic image understanding via global self-attention mechanisms. To leverage these complementary advantages, numerous semantic segmentation architectures have adopted strategies that integrate local and global perspectives [23].

One effective approach is the use of hybrid architectures, such as those fusing CNN and ViT paradigms. For instance, Wang et al. [24] proposed a framework that combines the strengths of CNN and ViT feature learning, employing a consistency-aware pseudo-label self-ensemble strategy to enhance semi-supervised semantic segmentation, showing promising results in medical image segmentation tasks. Beyond semi-supervised learning, Wang et al. [25] embedded CNN features into a ViT architecture, enabling the model to simultaneously capture global context and local multimodal information, thereby improving feature extraction in high-resolution remote sensing imagery. In a similar vein, Wang et al. [26] augmented MobileViT [27] with CNN enhancements to refine fine-grained region segmentation, applying it to semantic segmentation in wildfire scenarios.

Apart from hybrid architectures, many methods optimize the architectures themselves. For CNN-based models, the DeepLab series [28] introduced dilated convolutions to expand the receptive field, enhancing multi-scale perception through atrous convolutions at varying rates. HRNet [4] maintains high-resolution representations throughout the network via parallel multi-resolution branches, starting with a high-resolution branch (focusing on local details) and progressively adding lower-resolution branches (emphasizing global context), with repeated multi-scale fusions via exchange blocks. In contrast, ViT-based models emphasize on bolstering local detail capture. For example, Swin Transformer [22] employs localized self-attention to enforce focus on local regions, while alternating shifted windows in successive layers propagate information globally, achieving a balanced hierarchy. SegFormer [29] utilizes overlapping patch merging in a hierarchical Transformer encoder to extract multi-level features, streamlining global-local fusion without complex positional encodings.

## 2.2 Mamba-based segmentation networks

Mamba has emerged as a powerful architecture in natural language processing tasks [30], prompting researchers to explore its adaptation to the vision domain. Drawing inspiration from the Vision Transformer's [9] effective approach of dividing images into tokens, Vim [31] pioneers this effort by serializing image tokens and employing bidirectional Mamba blocks for compressed visual representation modeling—specifically, scanning from left to right (as shown in Fig. 1(a)) and from right to left in a Z-shaped pattern. Concurrently, VMamba [16] adopts a similar strategy but extends it to four-directional Z-shaped scanning: left-to-right (as in Fig. 1(a)), right-to-left, top-to-bottom, and bottom-to-top. These Mamba-based methods circumvent the quadratic complexity that constrains ViTs, making them more suitable for semantic segmentation of high-resolution images. Owing to their architectural superiority, simply connecting Vim or VMamba with high-performing decoders (such as UperNet) has yielded excellent results in semantic segmentation tasks [32].

Extensive efforts have also focused on refining Mamba-based architectures, with improvements primarily centered on optimizing scanning directions [33]. Samba [14] eschews multi-directional scanning in favor of a unidirectional sequence approach for efficient processing of high-resolution remote sensing images, achieving competitive segmentation results comparable to multi-directional strategies. More aggressively, the method proposed by Shi et al. [18] incorporates six directions, augmenting VMamba's four-directional scanning with two diagonal paths (as illustrated in Fig. 1(b)). While yielding improved performance, their approach incurs additional computational overhead from the increased directions. Similarly, Chen et al. [34] augmented Vim with a novel scanning strategy that shuffled tokens randomly. PlainMamba [17] deviates from the aforementioned cross-row jumping scans, arguing that continuity in scanning enhances spatial coherence and visual feature extraction. As depicted in Fig. 1(c), it employs an S-shaped serpentine scan. Beyond these, novel scanning directions include spiral [19] and tree-based [20] patterns. Despite the diversity in scanning directions across Mamba methods, the lack of fair quantitative comparisons leaves the improvements from these variants unclear [15]. Zhu et al. [15] addressed this in remote sensing image segmentation by proposing an experimental framework to evaluate various scanning directions and their combinations, demonstrating that differences in scanning directions and their stacking had minimal impact on performance.

In contrast to explorations of scanning directions, our proposed SwinMamba innovates at the mechanistic level, retaining VMamba-like four-directional scanning but, for the first time, effectively fusing local-global concepts with the Mamba architecture. Specifically, SwinMamba performs scanning on tokens within local regions in the early stages (as shown in Fig. 1(d)), facilitating inter-window information exchange through a shifted window-like mechanism, followed by global scanning in the later stages for holistic modeling. This approach enables the architecture to balance local and global information, thereby effectively supporting semantic segmentation tasks in remote sensing imagery.

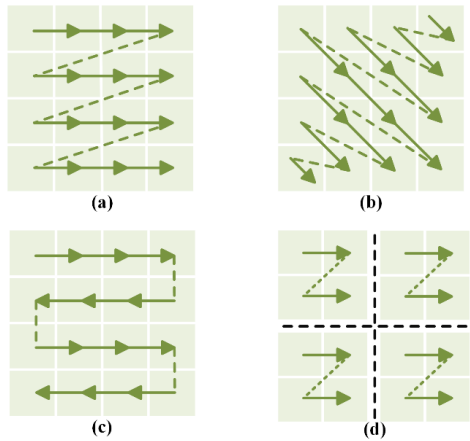

**Fig. 1.** Illustration of scanning strategies: (a) Horizontal scanning, (b) Diagonal scanning, (c) S-Shaped scanning, and (d) Our proposed region-based scanning.

## 3 SwinMamba

### 3.1 Overall architecture

The SwinMamba architecture is built upon an encoder-decoder framework, a structure widely adopted in networks for semantic segmentation tasks. The encoder comprises four stages: the first two stages consist of Local SwinMamba Blocks, which excel at capturing fine-grained local features, while the latter two stages employ Global SwinMamba Blocks, designed to integrate broader contextual information. Each stage utilizes progressive downsampling, transforming the input image from a size of $H \times W \times 3$, into feature maps with reduced spatial resolution and increased channel depth. Specifically, this process generates feature maps of sizes $\frac{H}{4} \times \frac{W}{4} \times C$, $\frac{H}{8} \times \frac{W}{8} \times 2C$, $\frac{H}{16} \times \frac{W}{16} \times 4C$, and ultimately $\frac{H}{32} \times \frac{W}{32} \times 8C$, as depicted in **Fig. 2**. This hierarchical design enables the model to effectively balance the extraction of detailed local structures and comprehensive global contexts, a critical factor for achieving high-accuracy semantic segmentation. The decoder leverages the state-of-the-art UperNet, producing the final segmentation result by aggregating multi-scale features.

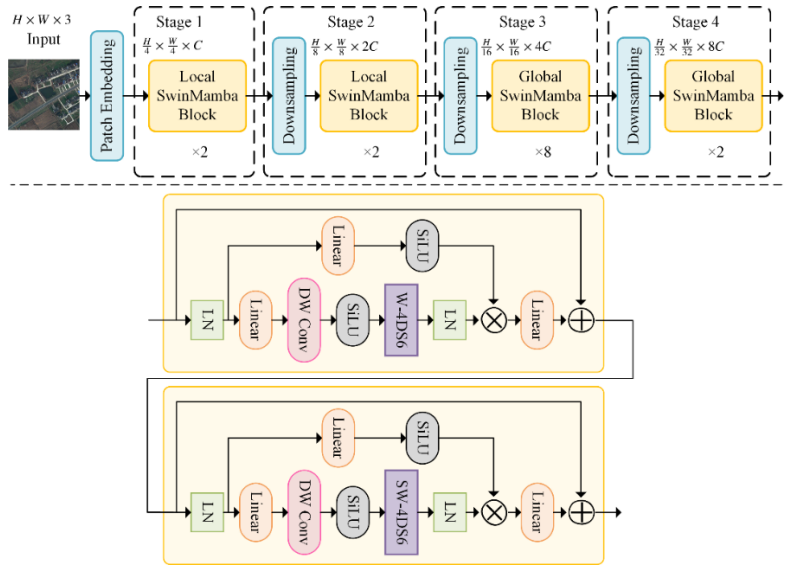

**Fig. 2.** The overall architecture of SwinMamba.

## 3.2 Local SwinMamba Block

Local SwinMamba Block employs a dual-configuration strategy to efficiently process spatial information. As illustrated in **Fig. 2**, odd-numbered blocks implement a Window-based Four-Directional Scanning S6 (W-4DS6) computation, partitioning the feature map into non-overlapping windows. These are immediately followed by even-numbered blocks that adopt a Shifted Window-based Four-Directional Scanning S6 (SW-4DS6) approach, where the windows are shifted to create overlapping regions with the previous windows. This alternating mechanism enhances the model's ability to capture fine-grained local details and cross-region dependencies.

In the W-4DS6 configuration (as shown in **Fig. 3 (a)**), the input feature map $F \in \mathbb{R}^{B \times C \times H \times W}$ is partitioned into non-overlapping rectangular windows of size $w \times w$.

$$F_{\text{win}} = \text{partition}(F, w) \in \mathbb{R}^{(B \cdot N) \times C \times w \times w} \tag{1}$$

where $N = \left(\frac{H}{w}\right) \times \left(\frac{W}{w}\right)$ represents the number of windows, and $B$ denotes the batch size. Within each window $F_{\text{win}}[i]$, a four-directional scanning process, covering horizontal and vertical directions, transforms the 2D feature map into 1D sequences:

$$S = \text{four\_direction\_scan}(F_{\text{win}}[i]) \tag{2}$$

These sequences are then processed through S6 computation, a hallmark of the Mamba architecture [12], which efficiently models dependencies with linear complexity:

$$Y = S6(S, \theta) \tag{3}$$

where $\theta$ denotes the model parameters.

In the SW-4DS6 configuration (as shown in **Fig. 3 (b)**), the window boundaries are shifted in even blocks, redefining the partitioning to cover adjacent regions. The feature map is first shifted by a distance $s$ (e.g., $s = 7$) along the spatial dimensions:

$$F_{\text{shift}} = \text{roll}(F, \text{shift} = (-s, -s), \text{dims} = (2, 3)) \tag{4}$$

The shifted feature map is then partitioned into windows and processed similarly:

$$F_{\text{win}}^{\text{shift}} = \text{partition}(F_{\text{shift}}, w) \tag{5}$$

$$S_{\text{shift}} = \text{four\_direction\_scan}(F_{\text{win}}^{\text{shift}}[i]) \tag{6}$$

$$Y_{\text{shift}} = S6(S_{\text{shift}}, \theta) \tag{7}$$

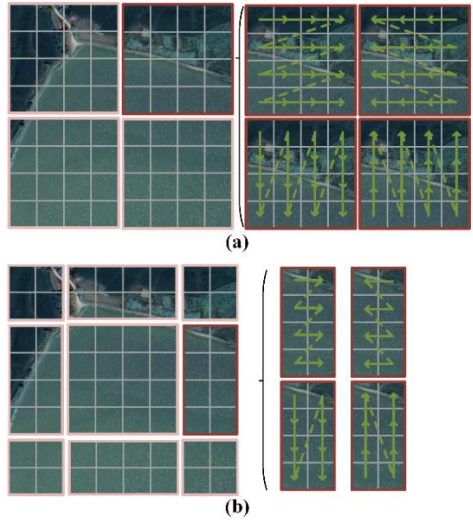

**Fig. 3.** (a) Window-based Four-Directional Scanning in odd blocks, (b) Shifted Window-based Four-Directional Scanning in even blocks.

This shifting scheme facilitates information exchange between windows, applying four-directional scanning and S6 computation to the newly partitioned regions.

Unlike Swin Transformer that optimizes self-attention for shifted windows through masking and concatenation, the S6 computation in the Mamba architecture inherently adapts to arbitrary rectangular shapes due to its sequential processing paradigm. This eliminates the need for additional optimization strategies, enabling seamless four-directional scanning across diverse window configurations.

## 3.3 Global SwinMamba Block

Following downsampling in the first two stages (i.e., Local SwinMamba Block), the input feature maps entering the third and fourth stages have shapes of $\frac{H}{16} \times \frac{W}{16} \times 4C$ and $\frac{H}{32} \times \frac{W}{32} \times 8C$, respectively. At these two stages, the spatial dimensions of the feature maps are significantly reduced, resulting in larger individual regions within the feature maps. Consequently, the Global SwinMamba Block abandons the window partitioning approach used in Local SwinMamba

Block and instead directly applies a global four-directional scanning strategy to the feature maps:

$$S_{\text{global}} = \text{global\_four\_direction\_scan}(F) \tag{8}$$

The resulting sequences are processed through the S6 module:

$$Y_{\text{global}} = \text{S6}(S_{\text{global}}, \theta) \tag{9}$$

This process transforms the feature maps into sequences scanned along horizontal and vertical directions, which are subsequently handled efficiently by the S6 computation to capture global dependencies and contextual relationships across the entire image.

## 4 Experiments

### 4.1 Datasets

To comprehensively evaluate the effectiveness of the SwinMamba architecture, we selected two classic remote sensing datasets for semantic segmentation: LoveDA [3] and ISPRS Potsdam. In our experiments, the dataset configurations were consistent with those used in other studies.

The LoveDA dataset comprises 2,522 training images, 1,669 validation images, and 1,796 test images. Each image is constituted of 1024×1024 pixels and three-channels RGB, and is featured with a spatial resolution of 0.3 meters. The dataset includes seven categories: background, building, road, water, barren, forest, and agricultural.

The ISPRS Potsdam dataset consists of 38 high-resolution remote sensing images, each with a spatial resolution of 5 cm and dimensions of 6000×6000 pixels. These images include near-infrared, red, green, and blue channels; however, in our experiments, only the red, green, and blue channels were utilized. The dataset encompasses six categories: impervious surface, building, low vegetation, tree, car, and clutter. Notably, the clutter category was excluded from the calculation of

validation metrics. Images with IDs 2_10, 2_11, 2_12, 3_10, 3_11, 3_12, 4_10, 4_11, 4_12, 5_10, 5_11, 5_12, 6_07, 6_08, 6_09, 6_10, 6_11, 6_12, 7_07, 7_08, 7_09, 7_10, 7_11, and 7_12 were used for training, while images with IDs 2_13, 2_14, 3_13, 3_14, 4_13, 4_14, 4_15, 5_13, 5_14, 5_15, 6_13, 6_14, 6_15, and 7_13 were employed for validation.

## 4.2 Pretraining

In the SwinMamba architecture, which follows an encoder-decoder paradigm, pretraining the encoder is a critical step to optimize its performance for downstream semantic segmentation of remote sensing imagery. Pretraining on large-scale datasets allows the model to learn robust, transferable representations that capture general visual patterns, thereby alleviating the challenges posed by limited labeled data in specialized domains like remote sensing. This is particularly important given the high-resolution and multi-spectral nature of remote sensing images, where direct training from scratch often leads to overfitting or suboptimal convergence. To this end, we conducted pretraining of the SwinMamba encoder on the ImageNet-1k dataset, a benchmark comprising approximately 1.28 million training images across 1,000 categories, widely recognized for its efficacy in fostering strong feature extractors. The classification performance metrics from the pretraining phase are presented in Table 1.

As anticipated, SwinMamba exhibited a marginal decline in image classification performance compared to the baseline VMamba model, which attained 82.6% top-1 accuracy. This slight decrement, amounting to 0.3%, was both expected and justifiable. VMamba's design leverages a global scanning mechanism in each module, which efficiently captures comprehensive token interactions across the entire image through a unified sequence conversion process. This global approach is inherently advantageous for image classification tasks, where holistic scene understanding often outweighs localized detail, and the balanced weighting of local and global information is less critical. In contrast, SwinMamba's scanning strategy deliberately prioritizes local token interactions via window-based partitioning and shifted mechanisms, which may not fully align with classification demands that favor immediate global context aggregation. However, this intentional design choice underscores SwinMamba's optimization for semantic segmentation, where focusing on local feature extraction, such as edges, textures, and fine-grained boundaries, is paramount for precise pixel-level predictions in complex remote sensing scenes.

**Table 1.** Classification performance of representative methods and SwinMamba in ImageNet-1k. The highest score is shown in bold.

| method | image size | top-1 acc |
|---|---|---|
| ViT-B | $384^2$ | 77.9 |
| Swin-t | $224^2$ | 81.3 |
| VMamba-t | $224^2$ | **82.6** |
| SwinMamba-t | $224^2$ | 82.3 |

## 4.3 Experimental setup

In this study, we compare our SwinMamba with several representative and state-of-the-art fully supervised semantic segmentation methods. These include CNN-based methods such as ConvNeXt [35] and ResNet [36], Vision Transformer (ViT)-based methods like CMTFNet [37], UNetformer [38], Swin Transformer [22], and Segformer [29], and the latest Mamba-based method, RS3Mamba [1], and VMamba [16]. The configurations of these baseline methods adhere to widely accepted optimal settings. We utilize the AdamW optimizer with a learning rate of 0.00006 and a batch size of 8 per GPU, conducting training over a total of 15,000 iterations. In the first two stages, the window size is set to 14, with a shift distance of 7. To assess the effectiveness of the proposed method, we employ the most commonly used metric mIoU. The segmentation results for CMTFNet, UNetformer, and RS3Mamba on the LoveDA and Potsdam datasets are as reported in [1] and .[39]

## 4.4 Results

Table 2 presents the segmentation performances on the LoveDA validation set. SwinMamba outperformed its baseline method VMamba-t by 1.06% in mIoU and notably outperformed Swin Transformer-t by 1.36%, demonstrating its enhanced ability to model complex spatial patterns. Similarly, Table 3 presents the segmentation performances on the ISPRS Potsdam validation set, which features high-resolution aerial imagery and fine-grained object boundaries. Here, SwinMamba outperformed VMamba-t by 0.33% and Swin Transformer-t by 0.59% in mIoU. The consistent performance improvement by SwinMamba across both datasets validated its effectiveness as a more advanced, accurate solution for semantic segmentation of remote sensing images.

**Table 2.** Segmentation performance of representative methods and SwinMamba in the LoveDA validation set.

| Decoder | Encoder | background | building | road | water | barren | forest | agricultural | mIoU |
|---|---|---|---|---|---|---|---|---|---|
| Upernet | ConvNeXt-t | **55.58** | 66.08 | 56.32 | 68.61 | 32.82 | 42.33 | **55.88** | 53.95 |
| Upernet | ResNet50 | 54.04 | 60.14 | 50.21 | 59.86 | 25.00 | 40.84 | 50.24 | 48.62 |
| Segformer-b0 | MixViT | 53.23 | 62.71 | 52.12 | 63.40 | 29.37 | 41.93 | 45.65 | 49.77 |
| CMTFNet | ResNet50 | 38.98 | 58.96 | 50.50 | 54.27 | 30.72 | 37.41 | 25.65 | 42.36 |
| UNetformer | ResNet18 | 37.61 | 52.78 | 51.89 | 63.47 | **39.84** | 34.23 | 11.44 | 41.61 |
| RS3Mamba | Mamba-t | 39.72 | 58.75 | 57.92 | 61.00 | 37.24 | 39.67 | 33.98 | 46.90 |
| Upernet | Swin-t | 54.61 | 65.37 | 55.98 | 69.79 | 31.69 | 44.14 | 52.19 | 53.40 |
| Upernet | VMamba-t | 55.18 | 65.93 | **57.9** | **71.97** | 32.38 | 40.78 | 51.77 | 53.70 |
| Upernet | SwinMamba-t(ours) | 55.05 | **66.03** | 57.62 | 70.60 | 33.91 | **44.19** | 55.28 | **54.76** |

**Table 3.** Segmentation performance of representative methods and SwinMamba in the Potsdam validation set.

| Decoder | Encoder | surface | building | vegetation | tree | car | mIoU |
|---|---|---|---|---|---|---|---|
| Upernet | ConvNeXt-t | 87.97 | 93.59 | 77.84 | 80.34 | 91.66 | 86.28 |
| Upernet | ResNet50 | 85.55 | 92.62 | 75.74 | 78.06 | 90.91 | 84.58 |
| Segformer-b0 | MixViT | 86.14 | 93.07 | 76.82 | 79.04 | 90.51 | 85.12 |
| CMTFNet | ResNet50 | 85.97 | 93.82 | 75.90 | 76.91 | 92.51 | 85.02 |
| UNetformer | ResNet18 | 85.87 | 93.32 | 74.43 | 75.55 | 91.56 | 84.15 |
| RS3Mamba | Mamba-t | 85.83 | 93.94 | 74.55 | 75.79 | 92.12 | 84.45 |
| Upernet | Swin-t | 88.23 | 94.34 | 77.92 | 80.18 | 91.65 | 86.46 |
| Upernet | VMamba-t | 88.21 | **94.72** | 77.96 | 80.13 | 92.60 | 86.72 |
| Upernet | SwinMamba-t(ours) | **88.58** | 94.69 | **78.57** | **80.51** | **92.92** | **87.05** |

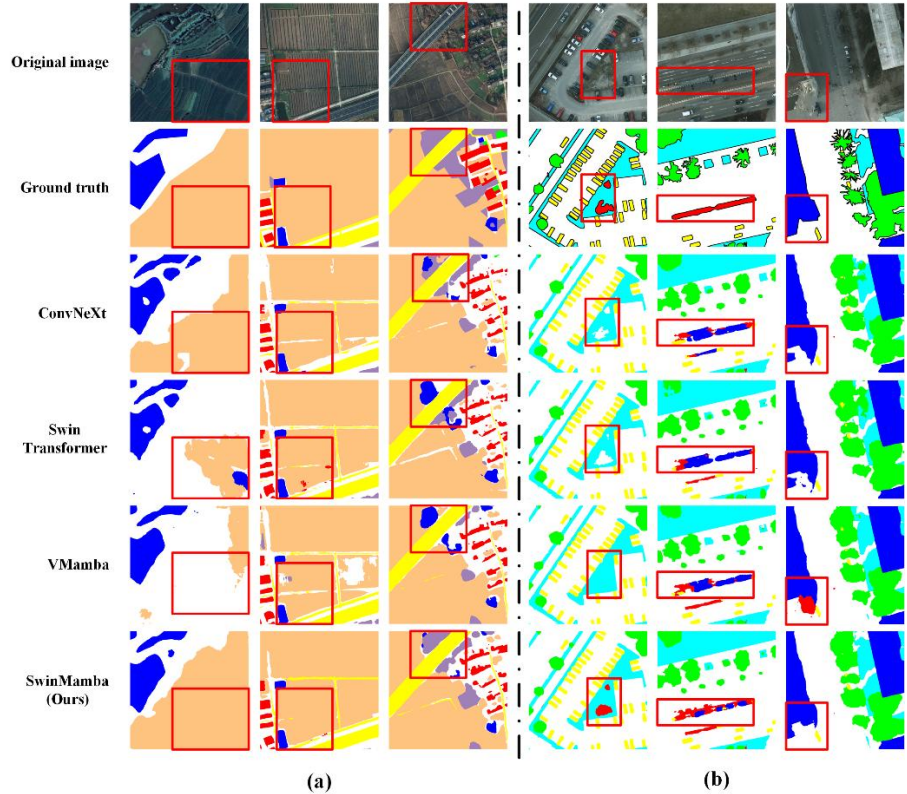

**Fig. 4.** Visual segmentation results of SwinMamba and other representative methods on (a) LoveDA, and (b) Potsdam datasets.

Furthermore, we conducted a visualization analysis of several example semantic segmentation results produced by SwinMamba and other representative methods on the LoveDA and Potsdam datasets, as illustrated in Fig. 4. SwinMamba demonstrated superior performance in mitigating errors such as false negatives and false positives compared to the baseline methods.

Specifically, in the first image (leftmost in Fig. 4(a)) from the LoveDA dataset, SwinMamba accurately captured most of the area representing agricultural category (depicted in orange [255, 195, 128]), as highlighted in the red-boxed region. However, the other tested networks exhibited varying degrees of missed detections for this category, with the baseline VMamba showing the most severe omissions. Similarly, the agricultural category in the red-boxed region of the second image was almost completely predicted by SwinMamba, outperforming the other tested methods. In the third image, the red-boxed barren category (depicted in purplish-gray [149, 129, 183]) was again best segmented by SwinMamba, while the other networks tended to misclassify this area as water (depicted in blue [0, 0, 255]).

On the Potsdam dataset, the red-boxed region in the first image (leftmost in Fig 4(b)) corresponds to the clutter category (depicted in red [255, 0, 0]), which was classified only by SwinMamba. Likewise, in the second image, the other networks tended to misclassify the clutter in the red-boxed area as building (depicted in blue [0, 0, 255]). In the third image, SwinMamba achieved more accurate boundary delineation for the building category, whereas the other methods exhibited various segmentation errors. Evidently, owing to its local-global architecture design, SwinMamba offers more refined local perception, leading to superior edge delineation and category classification compared to the baseline methods.

## 4.5 Ablation study

To validate the effectiveness of the scanning strategy in SwinMamba, where the first two stages employ local scanning and the latter two stages employ global scanning, we conducted a series of comparative experiments. These experiments evaluated three configurations: (1) all four stages using global scanning, (2) all four stages using local scanning, and (3) a hybrid configuration with local scanning in the first two stages and global scanning in the latter two. It is noteworthy that for the configuration using local scanning across all stages, the window size of 14 was infeasible in the final stage due to reduced spatial dimensions; thus, we adjusted it to 7. The experimental results, as shown in Table 4, indicate that our adopted hybrid scanning strategy achieved the highest mIoU on both the LoveDA and ISPRS Potsdam datasets. This was followed by the all-local scanning configuration, while the all-global scanning configuration performed the worst effectively. These findings underscore the advantage of stage-wise integration of local detail preservation and global contextual understanding.

**Table 4.** Segmentation performance (mIoU) using different local and global scanning strategies.

| Stage 1&2 | Stage 3&4 | LoveDA | Potsdam |
|---|---|---|---|
| Global | Global | 53.70 | 86.72 |
| Local | Local | 54.17 | 86.90 |
| Local | Global | **54.76** | **87.05** |

Additionally, we performed ablation experiments to assess the impact of window size and shift distance, comparing two settings: (1) a window size of 14 with a shift distance of 7, and (2) a window size of 7 with a shift distance of 3. As presented in Table 5, the influence of these parameters on overall performance was minimal. On the LoveDA dataset, the setting with a window size of 14 slightly outperformed the setting with a window size of 7 by 0.39% in mIoU, whereas on the Potsdam dataset, the setting with a window size of 7 outperformed the setting with a window size of 14 by 0.08%. Considering these marginal differences and computational efficiency, we ultimately selected the configuration with a window size of 14. These findings affirm the robustness of the hybrid scanning approach and its adaptability across different hyperparameter settings in remote sensing semantic segmentation tasks.

**Table 5.** Segmentation performance (mIoU) using different window size w and shift distance s.

| w | s | LoveDA | Potsdam |
|---|---|---|---|
| 7 | 3 | 54.37 | 87.13 |
| 14 | 7 | 54.76 | 87.05 |

## 5 Discussion and future work

In this study, SwinMamba has achieved substantial advancements in semantic segmentation of remote sensing imagery, particularly through its innovative integration of localized Mamba scanning with a shifted window mechanism. By employing local scanning in the early stages to capture fine-grained details and transitioning to global scanning in the later stages for comprehensive contextual fusion, SwinMamba effectively addresses the inherent multi-scale challenges in remote sensing data. Our empirical evaluations on the LoveDA and ISPRS Potsdam datasets reveal consistent improvements in mIoU, not only validating the efficacy of our hybrid scanning strategy but also highlighting its robustness across diverse datasets, positioning SwinMamba as a promising benchmark for future research in remote sensing applications.

Despite these accomplishments, several limitations merit discussion. First, the reliance on predefined window sizes in the local scanning stages may constrain adaptability to datasets with extreme resolution variations, necessitating manual adjustments to the sliding window size and shift distance in such scenarios, for instance, in ultra-high-resolution satellite imagery where object scales vary drastically. Additionally, while the local partitioning strategy significantly enhances performance in downstream semantic segmentation tasks, it mildly attenuates results in image classification, as evidenced in Table 1. Mitigating the impact of local region partitioning on global image understanding remains an area for consideration, particularly in scenarios where holistic scene comprehension is paramount.

Looking ahead, future work could explore several avenues to enhance SwinMamba's capabilities. A promising direction is efficiency optimization. Our region partitioning strategy generates sub-regions where the majority exhibit consistent token lengths, enabling parallelized computation across these uniform segments, which could substantially improve efficiency compared to VMamba's global scanning approach that precludes such parallelism. Although our current implementation does not fully capitalize on this potential in engineering terms, future efforts could incorporate GPU-accelerated parallelization techniques, such as distributed tensor processing, to realize these benefits and reduce inference time in large-scale remote sensing pipelines. Furthermore, as noted in Section 2.2, the shifted window partitioning produces sub-regions with varying token lengths; while Mamba's S6 computation accommodates arbitrary lengths without additional adaptations, we could draw inspiration from Swin Transformer's token concatenation and masking strategies to standardize sequence lengths, developing S6-specific masking mechanisms to further enhance parallelization and promote more uniform and efficient computations. Moreover, future work could consider adaptive mechanisms for window size and shift distance to eliminate these as external hyperparameters. Finally, deploying SwinMamba in real-time systems, such as edge devices for disaster monitoring, would necessitate model compression techniques like pruning or quantization to ensure scalability while preserving segmentation accuracy. These explorations will not only refine SwinMamba but also propel broader progress in the field of remote sensing semantic segmentation.

## 6 Conclusion

In this paper, we present SwinMamba, which integrates localized Mamba-style scanning within shifted windows with a global receptive field to significantly enhance both local and global feature extraction for semantic segmentation of remote sensing images. Experiments conducted on the LoveDA and ISPRS Potsdam datasets demonstrated its superior performance compared to several state-of-the-art methods, validating its effectiveness.

While SwinMamba excels in multi-scale feature fusion, its predefined window sizes may limit adaptability to extreme resolutions, and the shifted partitioning could introduce minor computational overhead. These trade-offs highlight opportunities for further refinement. Future work can prioritize efficiency enhancements through approaches such as GPU-accelerated parallelization of uniform token-length sub-regions and S6-specific masking to standardize sequences. In addition, future work can extend the model toward multimodal data fusion and adaptive window mechanisms, further broadening its applicability to real-time remote sensing tasks while maintain its performance in segmentation accuracy.

## Declaration of competing interest

The authors declare that they have no known competing financial interests or personal relationships that could have appeared to influence the work reported in this paper.

# Acknowledgments

This work was supported by the Xi'an Jiaotong-Liverpool University Postgraduate Research Scholarship under Grant FOS2210JJ03.